\documentclass[letterpaper, 10pt, conference]{ieeeconf}

\IEEEoverridecommandlockouts
\usepackage[utf8]{inputenc}
\usepackage{amssymb,amsfonts,amsmath,amscd}
\usepackage{bm}
\usepackage[pdftex]{graphicx}
\usepackage{url}
\usepackage{cite}
\usepackage[usenames,dvipsnames]{color}
\usepackage{subcaption}
\usepackage{caption}
\usepackage{accents}
\usepackage{tabularx}
\usepackage{tikz}
\usepackage{stackengine}
\usepackage{hyperref}

\let\labelindent\relax % required to allow enumitem to exist with ieeeconf
\usepackage{enumitem}

\usepackage{mathtools}
\usepackage[english]{babel}
\usepackage{siunitx}
\usepackage[export]{adjustbox}

\usepackage{microtype}

\usepackage{booktabs}
\usepackage{makecell}
\usepackage{multirow}

\definecolor{pltred}{rgb}{0.839, 0.153, 0.157}

\newcommand{\barf}{\skew{5}{\bar}{\bm{f}}}

\DeclareMathOperator{\diag}{\mathrm{diag}}

\newcommand{\R}{\mathbb{R}}

\title{\LARGE\textbf{Robust Single-Point Pushing with Force Feedback}}
\author{Adam Heins and Angela P. Schoellig%
  \thanks{The authors are with the Learning Systems and Robotics Lab (\href{www.learnsyslab.org}{www.learnsyslab.org}) at the Technical University of Munich, Germany, and the University of Toronto Institute for Aerospace Studies, Canada. They are also affiliated with the University of Toronto Robotics Institute, the Munich Institute of Robotics and Machine Intelligence (MIRMI), and the Vector Institute for Artificial Intelligence. E-mail: adam.heins@robotics.utias.utoronto.ca, angela.schoellig@tum.de}
}

\begin{document}

\maketitle

\begin{abstract}
  We present the first controller for quasistatic robotic planar pushing with
  single-point contact using \emph{only} force feedback. We
  consider a mobile robot equipped with a force-torque sensor to measure the
  force at the contact point with the pushed object (the ``slider''). The parameters of the
  slider are not known to the controller, nor is feedback on the
  slider's pose. We assume that the global position of the contact point is
  always known and that the approximate initial position of the slider is
  provided. We focus specifically on the case when it is desired to push the
  slider along a straight line. Simulations and real-world experiments show
  that our controller yields stable pushes that are robust to a wide range of
  slider parameters and state perturbations.
\end{abstract}

\section{Introduction}

Pushing is a nonprehensile manipulation primitive that allows robots to
move objects without grasping them, which is useful for objects that are too
heavy, cumbersome, or delicate to be reliably grasped. In this work we
investigate robotic planar pushing with single-point contact using only force
feedback. The pusher is a mobile robot equipped with a force-torque (FT) sensor
to measure the contact force between the robot and the pushed object (``the
slider''). The parameters of the slider are not known---this includes its
geometry and inertial parameters like mass and center of mass (CoM), but we
assume the slider's shape is convex. It is assumed that an approximate initial
position of the slider is known but that online feedback of its pose is not
available---the only measurement of the slider is through the contact force
with the pusher. We assume that the global position of the pusher is known at
all times (i.e., the robot can be localized). Finally, we assume that all
motion is quasistatic.

We focus specifically on pushing the slider along a desired straight-line path,
for which we found our controller to be particularly suited. We envision such an
approach being useful for pushing unknown objects between distant waypoints,
where reliable localization of the object is not available. For example,
consider pushing objects through long, straight hallways within warehouses or
factories.

\begin{figure}[t]
  \centering
    \includegraphics[width=\columnwidth]{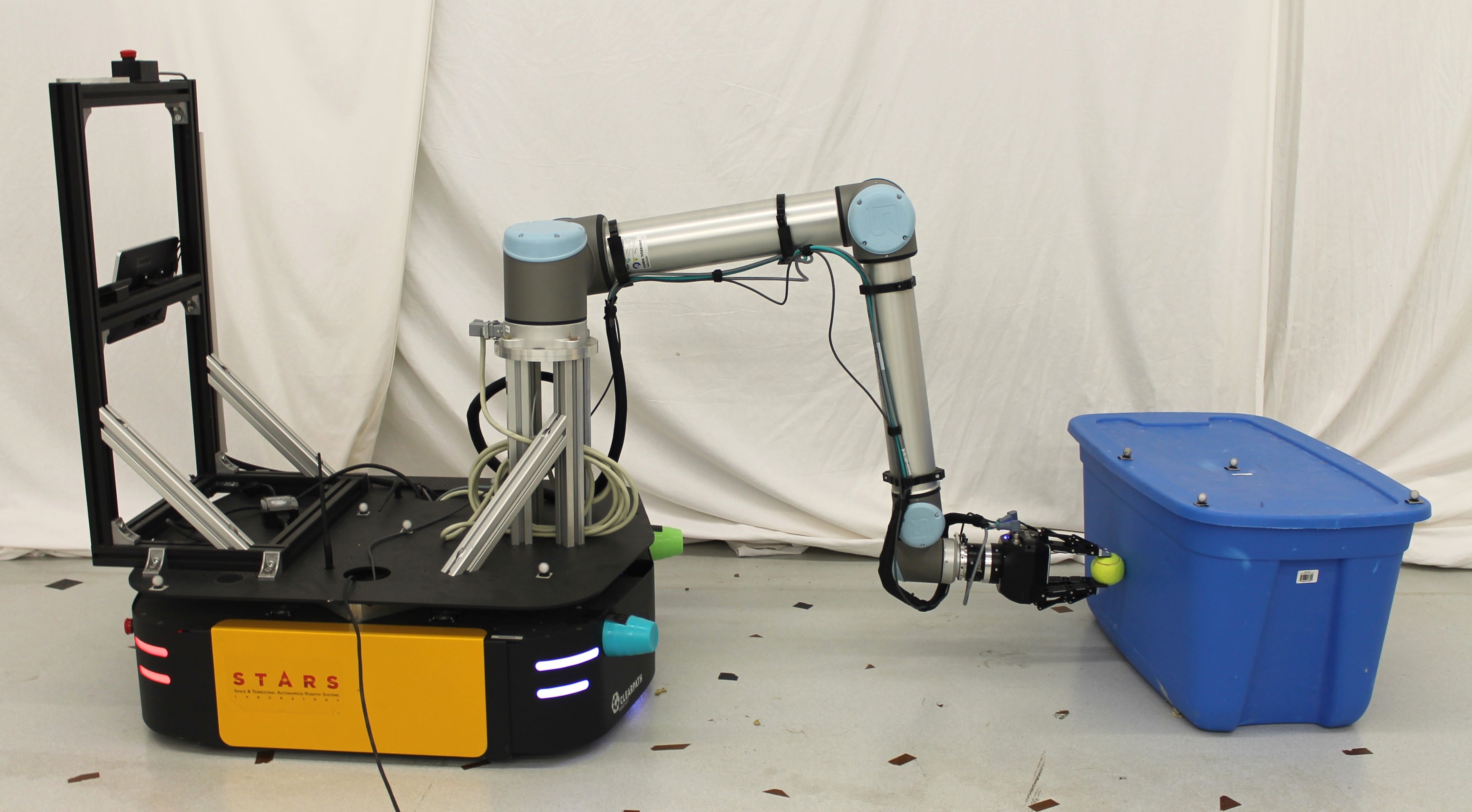}
    \caption{Our robot pushing a blue box across the floor using
      single-point contact. The contact force is measured by a force-torque
      sensor in the robot's wrist, but no other measurements of the object are
      provided.}
  \label{fig:eyecandy}
\end{figure}

The main contribution of this work is to present the first controller for
pushing objects with single-point contact based only on force feedback. We show
that it successfully pushes objects along a desired straight-line path with
single-point contact. We demonstrate the robustness of the controller by
simulating pushes using a wide variety of slider parameters and initial states.
We also present real hardware experiments in which a mobile manipulator
successfully pushes different objects across a room (see
Fig.~\ref{fig:eyecandy}). Notably, we do not assume that sufficient friction is
available to prevent slip at the contact point. Indeed, we will see that
slipping is a natural part of the behaviour of our controller and does not
necessarily lead to task failure.

We first briefly described a preliminary version of this controller
in~\cite{heins2021mobile}. In the current work we refine the control law, add a
term to track a desired line, provide an analysis of its robustness in
simulation, and perform more numerous and challenging real-world experiments.

\section{Related Work}

Research on robotic pushing began with Mason~\cite{mason1986mechanics}.
The approaches that followed were typically either open-loop
planning methods that rely on multi-point contact with a fence for
stability~\cite{lynch1996stable,akella1998posing} or feedback-based approaches based on
vision~\cite{emery2001behavior,igarashi2010a} or tactile
sensing~\cite{okawa1992control,lynch1992manipulation}. Tactile sensing is the
most similar to our work, though we assume only a single contact force vector
is available, rather than the contact angle and normal that a tactile sensor
provides.

An FT sensor is used with a fence to orient polygonal parts using a
sequence of one-dimensional pushes in~\cite{rusaw1999part}, which was shown to
require less pushes than the best sensorless
alternative~\cite{akella2000parts}. Another use of an FT sensor was
in~\cite{ruizugalde2011fast}, where FT measurements are used to detect slip
while pushing. In constrast, we do not detect slip; rather, our closed-loop
dynamics are stable despite (unmeasured) slip.

More recent work has turned to learned-based approaches to model the
complicated pushing dynamics arising from uncertain friction
distributions and object parameters~\cite{bauza2018a,li2018push}. Another line of
work~\cite{hogan2020feedback,agboh2020pushing} uses model predictive control
(MPC) for fast online replanning. These methods are powerful but typically
assume at least visual feedback is available (to localize the slider) and
require either considerable training data for learning or an existing model of
the slider. A survey on robotic pushing can be found in~\cite{stuber2020lets}.
None of these methods are based \emph{only} on force feedback.

\section{Methodology}

We first describe the model of the slider used for simulation and then
present our controller for pushing the slider based on force feedback.

\subsection{Model of Quasistatic Pushing}

\begin{figure}[t]
  \centering
    \includegraphics[width=\columnwidth]{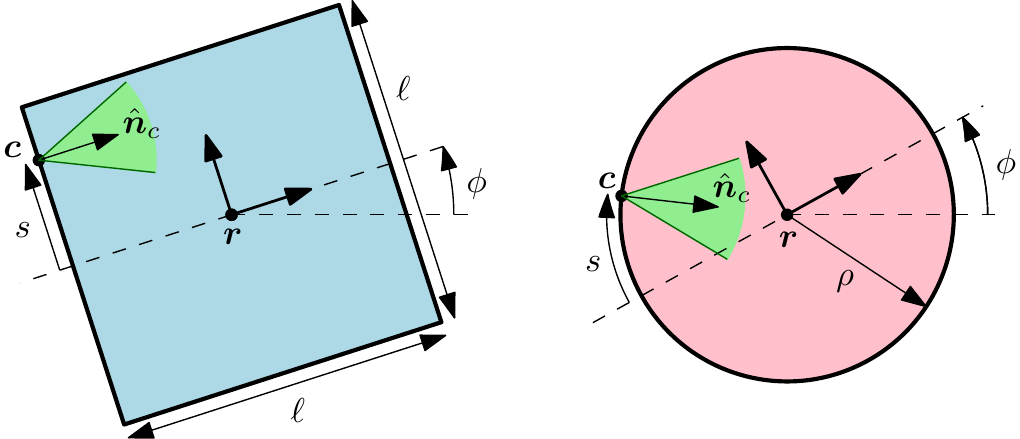}
    \caption{Examples of a square and a circular slider, located at
      position~$\bm{r}$ and orientation~$\phi$ in the global frame. The contact
      with the pusher is located at point~$\bm{c}$, which is distance~$s$
      along the slider's edge. The contact force must lie in the friction cone at
      the contact point (shown in green).}
  \label{fig:setup}
\end{figure}

We use a model of quasistatic planar pushing equivalent to that presented
in~\cite{lynch1992manipulation}, which we briefly review here.
Let~$\bm{p}=[f_x,f_y,\tau]^T$ be the generalized force acting on the slider,
where~$(f_x,f_y)$ is the force and~$\tau$ is the torque. Under the
ellipsoidal limit surface model, this force must lie within the ellipsoid
defined by
\begin{equation}\label{eq:limit_surface_ellipsoid}
  \bm{p}^T\bm{M}\bm{p}\leq 1,
\end{equation}
where~$\bm{M}=\diag(f_{\max}^{-2},f_{\max}^{-2},\tau_{\max}^{-2})$ represents
the maximum frictional load between the slider and the support plane
(i.e., the ground). When the object is moving, \eqref{eq:limit_surface_ellipsoid} holds
with equality. For simplicity we will assume uniform support friction, which
means that the center of friction corresponds to the slider's CoM
projected onto the support plane.
If we assume that the slider's pressure distribution is concentrated at a
particular distance~$d$ from the CoM, then~$\tau_{\max} = d\cdot f_{\max}$.
When the pressure distribution is uniform,
then~$d=(\int_{A}\|\bm{a}\|dA)/(\int_{A}dA)$,
where~$A$ is the support area,~$dA$ is a differential element of area
of~$A$, and $\bm{a}$ is the position of~$dA$.

The system state is~$\bm{x}=[x,y,\phi,s]^T$, where~$\bm{r}=[x,y]^T$ is the slider's
position, $\phi$ is its orientation, and~$s$ is the distance of the contact
point~$\bm{c}$ along the slider's edge (see Fig.~\ref{fig:setup}). The slider's
generalized velocity about the CoM and expressed in the body frame
is~$\bm{\varpi}=[v_x,v_y,\omega]^T$, with linear velocity~$(v_x,v_y)$ and angular velocity~$\omega$. Under quasistatic pushing, $\bm{\varpi}$
is in the direction normal to the boundary
of~\eqref{eq:limit_surface_ellipsoid}. Let~$\bm{v}_o\in\R^2$ be the slider's
velocity at the contact point and~$\bm{v}_p\in\R^2$ be the pusher's velocity;
these are equal if the contact is sticking, but the magnitude of~$\bm{v}_p$
is larger during slipping. We have the
relationship~$\bm{v}_o=\bm{W}_c^T\bm{\varpi}$, where
\begin{equation*}
  \bm{W}_c^T = \begin{bmatrix} 1 & 0 & -y_c \\ 0 & 1 & x_c \end{bmatrix},
\end{equation*}
for contact point~$\bm{c}=[x_c,y_c]^T$ expressed in the body frame.

We can conveniently express the slider's equations of motion as the
solution to the quadratic program
\begin{equation}\label{eq:qp_equations_of_motion}
  \begin{aligned}
    \min_{\bm{\varpi},\alpha,\bm{\eta}} &\quad \alpha^2 \\
    \text{subject to} &\quad \bm{\varpi} = \bm{M}\bm{W}_c\bm{\eta} \\
                      &\quad \bm{v}_p = \bm{W}_c^T\bm{\varpi} + \alpha\hat{\bm{n}}_c^{\perp} \\
                      &\quad \bm{\eta} \in \mathcal{FC},
  \end{aligned}
\end{equation}
where~$\alpha$ is the contact's velocity along the current edge of the slider (i.e., the slip velocity),
$\bm{\eta}$ is a vector parallel to the contact force, $\hat{\bm{n}}_c^{\perp}$
is a unit vector perpendicular to the contact normal, and~$\mathcal{FC}=\{\bm{f}\in\R^2\mid|\hat{\bm{n}}_c^\perp\cdot\bm{f}| \leq \mu_c\hat{\bm{n}}_c\cdot\bm{f}\}$ is the
friction cone at the contact with friction coefficient~$\mu_c$. If desired, we
can obtain the actual contact force~$\bm{f}$ by
substituting~$\bm{p}=\bm{W}_c\bm{\eta}$ into the boundary of the limit
surface~\eqref{eq:limit_surface_ellipsoid} to obtain
\begin{equation*}
  \bm{f} = (\bm{\eta}^T\bm{W}_c^T\bm{M}\bm{W}_c\bm{\eta})^{-1/2}\bm{\eta}.
\end{equation*}
Intuitively,~\eqref{eq:qp_equations_of_motion}
tries to find the object velocity~$\bm{\varpi}$ which corresponds to a feasible
contact force and produces as little slip as possible. The system's contact
modes (sticking, sliding left, sliding right) correspond to particular sets of
active constraints
of~\eqref{eq:qp_equations_of_motion}. The pusher's contact velocity~$\bm{v}_p$
is given as the input to the system. If~$\bm{v}_p$ is pulling away from the
object (i.e.,~$\hat{\bm{n}}_c\cdot\bm{v}_p<0$),
then~\eqref{eq:qp_equations_of_motion} is infeasible. Assuming the pusher and
slider maintain contact, we can simulate the entire system forward in time by
integrating~$\bm{\varpi}$ and~$\alpha$.
The formulation~\eqref{eq:qp_equations_of_motion} produces the same results as
the analytical equations in~\cite{lynch1992manipulation}, but the
correspondence between contact modes and active constraints may aid the
intuition.

\subsection{Pushing Controller}

\begin{figure}[t]
  \centering
    \includegraphics[width=0.9\columnwidth]{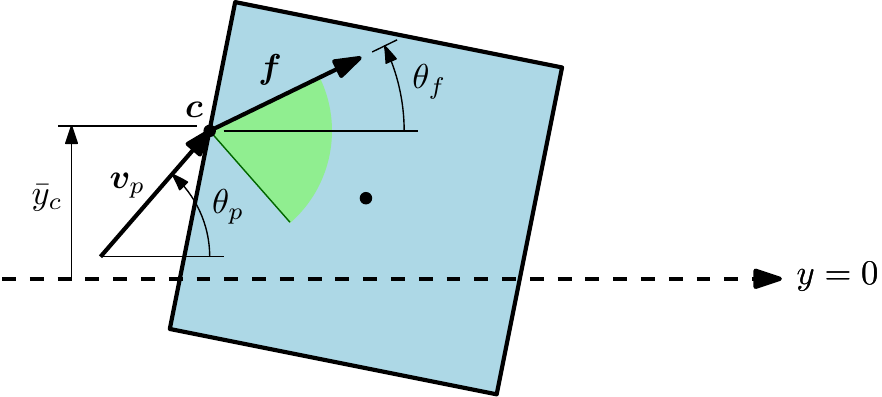}
    \caption{Example of our the pushing controller. The goal is to push the box
    along the line~$y=0$ by pushing with velocity~$\bm{v}_p$ at the contact
    point~$\bm{c}$. The pushing angle~$\theta_p$ is proportional to the lateral
    offset~$\bar{y}_c$ and measured force angle~$\theta_f$. In this example,
    the commanded~$\bm{v}_p$ will eventually rotate the object so that the
    contact force points back toward the desired path along~$y=0$.
    % The controller steers the slider so as to converge to a configuration where both~$\bm{c}$ and the CoM lie along the line~$y=0$.
    Depending on the contact friction
    coefficient~$\mu_c$, the contact point is free to slip along the object's
    edge over the course of a trajectory.}
  \label{fig:controller}
\end{figure}

We now turn our attention to generating the pusher velocity~$\bm{v}_p$ given
measurements of the contact force~$\bm{f}$. Let~$\bar{\bm{v}}_p$
and~$\barf$ be these same quantities but expressed in the global frame;
indeed, since we do not assume to know the orientation of the slider, we cannot
work in the local frame. Let us express these quantities in polar form as
\begin{align*}
  \bar{\bm{v}}_p &= \|\bar{\bm{v}}_p\|\cdot\begin{bmatrix} \cos\theta_p \\ \sin\theta_p \end{bmatrix}, &
  \barf &= \|\barf\|\cdot\begin{bmatrix} \cos\theta_f \\ \sin\theta_f \end{bmatrix}.
\end{align*}
We set the pushing speed~$\|\bar{\bm{v}}_p\|$ constant but control the
pushing angle~$\theta_p$. Our goal is to track a given straight-line path,
which we will assume without loss of generality to be along the $x$-axis of the
global frame (i.e., the line~$y=0$ with~$x$ increasing). Given an approximate
initial position of the slider, we assume that the robot can be positioned so that
it starts in contact with the slider and is approximately aligned with the desired
pushing direction.
Our control law is simply
\begin{equation}\label{eq:ctrl_law}
  \theta_p = (k_f+1)\theta_f + k_y\bar{y}_c,
\end{equation}
where~$k_f,k_y>0$ are tunable gains and~$\bar{y}_c$ is the lateral deviation
of the contact point from the desired path (see Fig.~\ref{fig:controller}). The
first term steers toward a stable pushing direction and the second term steers
toward the desired path. Ultimately, the controller converges to a
configuration where the contact force lies along the
desired path. Notice that the controller does not depend explicitly on any
slider parameters, and can thus be used to push a variety of unknown objects. In
particular, we do not require knowledge of the support friction, pressure
distribution, or contact friction, which are often uncertain and subject to
change. Depending on the contact friction coefficient~$\mu_c$, the contact
point may slip or stick along the edge of the object over the course of a
successful push, which is not a problem for our controller.

\section{Simulations}

We first validate our controller in simulation with a square and circular slider,
as shown in Fig.~\ref{fig:setup}, using the equations of
motion~\eqref{eq:qp_equations_of_motion} and our control
law~\eqref{eq:ctrl_law}. The square has side length~$\ell=\SI{1}{m}$ and the
circle has radius~$\rho=\SI{0.5}{m}$; the CoMs are located at the
centroids. For each slider, we assess the robustness of our controller by
running simulations with different combinations of initial
state~$\bm{x}_0=[0,y_0,\phi_0,s_0]^T$, contact friction, and pressure
distribution (encoded in~$\tau_{\max}$), as listed in
Table~\ref{tab:parameters}. We use pusher
speed~$\|\bar{\bm{v}}_p\|=\SI{0.1}{m/s}$, controller gains~$k_f=0.1$
and~$k_y=0.01$, and~$f_{\max}=\SI{1}{N}$. The simulation timestep
is~\SI{10}{ms}.

\begin{table}[t]
  \caption{Initial states and parameters used for simulation of the square and
    circular sliders. Every combination of states and parameters is used, for a
    total of~$3^5=243$ combinations per slider. The values of~$\tau_{\max}$
    depend on the slider shape, with~$\bar{\tau}$ computed assuming a uniform
    pressure distribution and~$\hat{\tau}$ computed assuming the pressure is
    concentrated at maximum distance from the CoM.}
  \centering
  \begin{tabular}{l c c c}
    \toprule
    Parameter & Symbol & Values & Unit \\
    \midrule
    Initial lateral offset & $y_0$ & $-40$, $0$, $40$ & \si{cm} \\
    \midrule
    Initial contact offset & $s_0$ & $-40$, $0$, $40$ & \si{cm} \\
    \midrule
    Initial orientation & $\phi_0$ & $-\pi/8$, $0$, $\pi/8$ & \si{rad} \\
    \midrule
    Contact friction & $\mu_c$ & $0$, $0.5$, $1.0$ & $-$ \\
    \midrule
    Max. torsional load & $\tau_{\max}$ & $0.1\bar{\tau}$, $\bar{\tau}$, $\hat{\tau}$ & \si{Nm} \\
    \bottomrule
  \end{tabular}
  \label{tab:parameters}
\end{table}

\begin{figure}[t]
  \centering
    \includegraphics[width=\columnwidth]{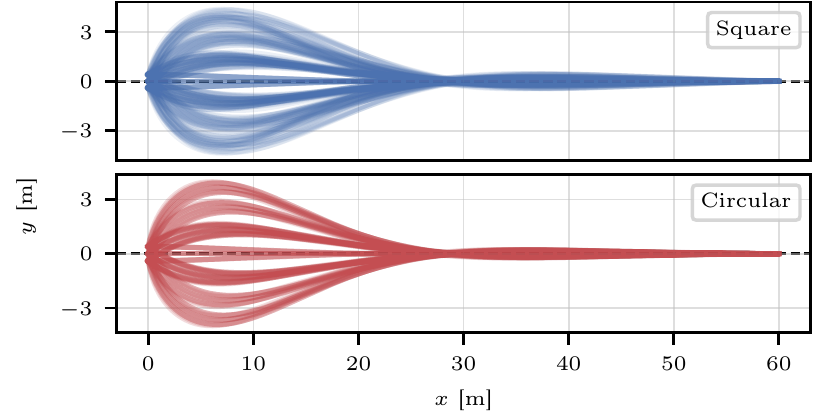}
    \caption{Simulated trajectories for all $243$ combinations of parameters
      given in Table~\ref{tab:parameters} for the square and circular sliders
      shown in Fig.~\ref{fig:setup}. Each trajectory has a
      duration of~\SI{10}{min}. All trajectories converge to the desired
      path~$y=0$ using our control law with the same set of gains~$k_f=0.1$
      and~$k_y=0.01$.}
  \label{fig:simulate_many}
\end{figure}

The position trajectories for each of the~$3^5=243$ parameter combinations per
slider are shown in Fig.~\ref{fig:simulate_many}. Our controller successfully
steers both sliders to the desired path along the positive $x$-axis for every
parameter combination with the same controller gains. While~$k_y$ could be
increased to reduce the deviation from the desired path, we found that a
larger~$k_y$ was not stable for all of these parameter combinations.

\begin{figure}[t]
  \centering
    \includegraphics[width=\columnwidth]{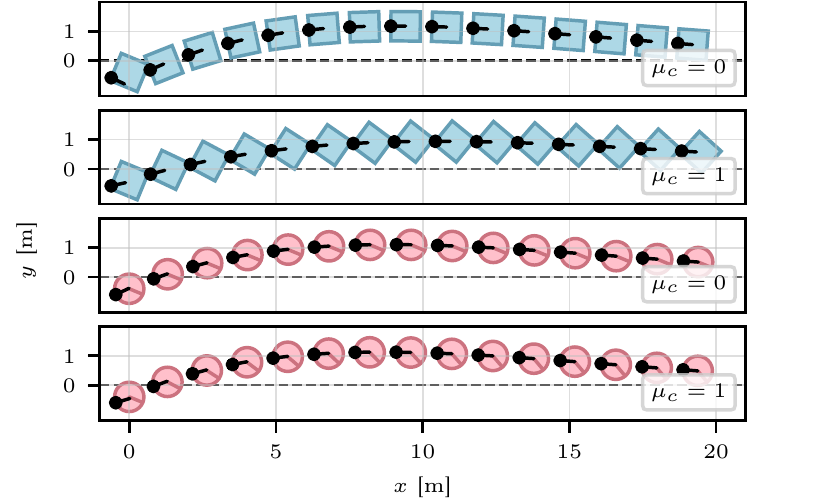}
    \caption{Samples of simulated trajectories with
      initial state~$(x_0,y_0,s_0,\phi_0)=(0,\SI{-40}{cm},\SI{-40}{cm},-\pi/8)$ and a uniform pressure
      distribution. Results are shown for each slider with low and high
      contact friction. Contact point and pushing direction are shown in black.
      The orientation of the circular slider is indicated by the red line between
      its center and edge, which is aligned with the body frame's $x$-axis.}
  \label{fig:simulate_few}
\end{figure}

Four sample trajectories are shown in Fig.~\ref{fig:simulate_few}. In
particular, notice the difference between the two trajectories of the square
slider. The first has zero contact friction ($\mu_c=0$), and we see that the
contact point quickly slips toward the center of the square's edge. In
contrast, the second trajectory has high contact friction ($\mu_c=1$): the
contact point does not slip and a stable push is achieved with a large
angle between the contact normal and pushing direction.

\section{Hardware Experiments}

\begin{figure}[t]
  \centering
    \includegraphics[width=0.8\columnwidth]{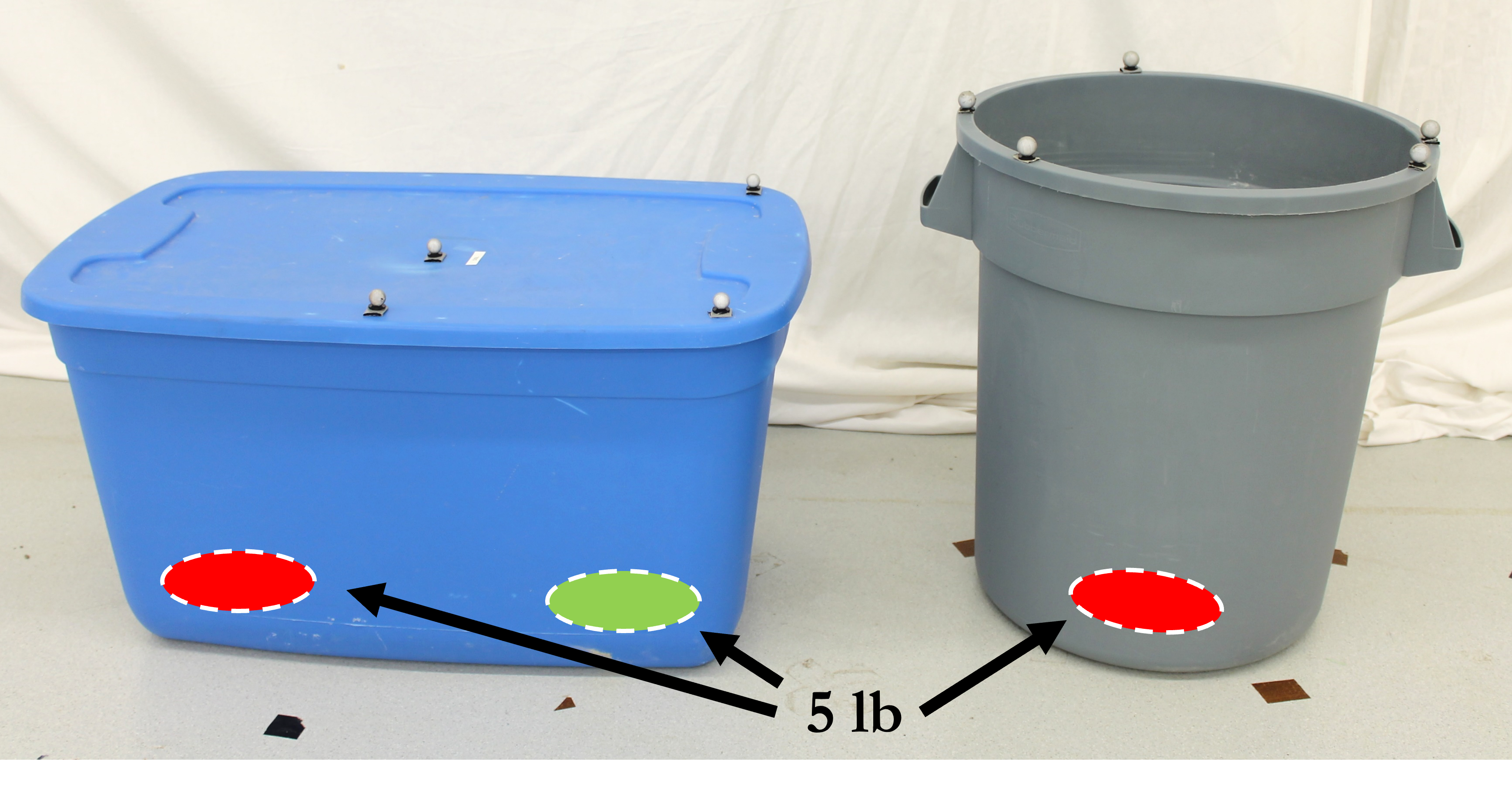}
    \caption{The ``box'' and ``barrel'' sliders used for real-world
      experiments. Each is empty except for~\SI{5}{lb} weights
      located approximately at the colored circles. The red weights are always
      present, but we add or remove the green weight to vary the mass and
      pressure distribution of the box.
      Motion capture is used to record the sliders' trajectories, but is not
      used for control.}
  \label{fig:objects}
\end{figure}

We now demonstrate our controller in real-world experiments.
The robot used for pushing is a mobile manipulator consisting of a UR10 arm
mounted on a Ridgeback omnidirectional base (see Fig.~\ref{fig:eyecandy}). The
arm's wrist is equipped with a Robotiq FT 300 force-torque sensor. Since we are
only concerned with pushing in the $x$-$y$ plane, we fix the joint angles of
the arm and the orientation of the base and only control the linear velocity of
the base by commanding~$\bar{\bm{v}}_p$. The base is localized using a Vicon
motion capture system, which is also used to record the trajectories of the
sliders.

We test our controller's ability to push a box and a barrel (shown in
Fig.~\ref{fig:objects}) across the floor, each of which contains~$\SI{5}{lb}$
weights. We vary the amount of weight (either~$5$ or~\SI{10}{lb}) in the box to
provide additional variety in the slider mass and pressure distribution. We assume
that the height of the contact point is such that the sliders do not tip over.

We continue to use~$\|\bar{\bm{v}}_p\|=\SI{0.1}{m/s}$ and~$k_f=0.1$, but
we increase~$k_y$ to~$0.3$. A lower value of~$k_y$ was required in simulation
so that the controller was stable for \emph{all} combinations of parameters,
but for these real-world experiments we found that a higher~$k_y$ is more
effective for tracking the desired path~$y=0$. In general, the controller gains
can be tuned to give better tracking performance when the set of possible slider
parameters is smaller.

\begin{figure}[t]
  \centering
    \includegraphics[width=\columnwidth]{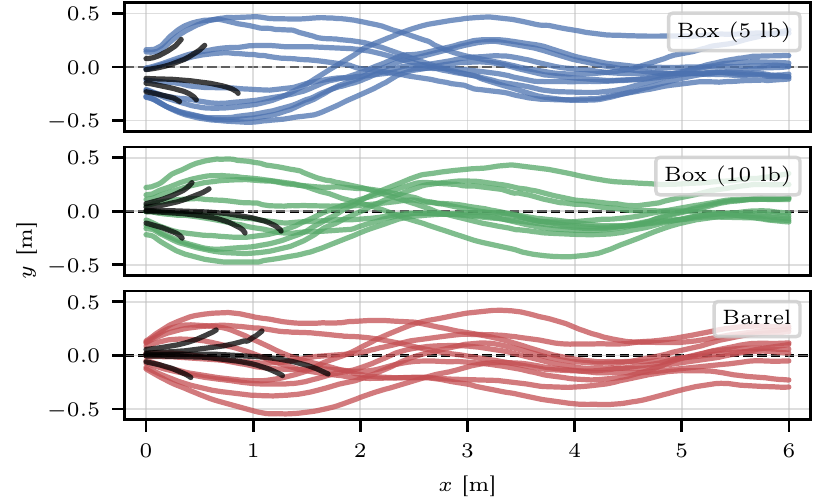}
    \caption{Position trajectories for real sliders pushed starting from various
    lateral offsets. Ten trajectories using our pushing control law are shown
    for each slider (in color). We compare against an open-loop controller,
    which just commands the robot to move forward with constant
    speed~$\|\bar{\bm{v}}_p\|$. Five open-loop trajectories are shown for
    each slider (in black). The open-loop trajectories end once contact between
    the pusher and slider is lost. All open-loop trajectories fail within~\SI{2}{m}, whereas our
    controller is able to push the objects across the full length of the room.}
  \label{fig:experimental_results}
\end{figure}

The experimental results are shown in Fig.~\ref{fig:experimental_results}. Ten
trajectories using our pushing control law are shown for each slider. We
compare against an open-loop controller, which just commands the robot to move
forward with constant speed~$\|\bar{\bm{v}}_p\|$. Five open-loop trajectories are
shown for each slider. Open-loop pushing with single-point contact is
unstable, and indeed we see that the open-loop trajectories quickly fail. In
contrast, our closed-loop pushing controller successfully pushes the sliders
across the full~\SI{6}{m} length of the room.

It should be noted that the trajectories do not
converge perfectly to the desired path~$y=0$, at least not within the
available~\SI{6}{m} distance. This is expected in the real world, as the slider
is constantly perturbed by imperfections on the surface of the ground as
it slides, which must then be corrected by the controller. Indeed, as can be
partially seen in Fig.\ref{fig:eyecandy}, the floor of the room has various
pieces of tape and other markings which change the surface friction properties
as the object slides. Regardless, in Fig.~\ref{fig:experimental_results} we see
that the controller keeps the slider within a corridor of
approximately~\SI{1}{m} width around~$y=0$, even with considerable initial
lateral offsets between pusher and slider.

\section{Conclusion}

We presented a control law for quasistatic robotic planar pushing with
single-point contact using only force feedback, which does not require known
slider parameters or pose feedback, and we demonstrated its robustness in simulated and
real-world experiments. Subsequent work will further investigate the
theoretical properties of the controller and apply it to more challenging
environments, such as those with obstacles.

\bibliographystyle{IEEEtran}
\bibliography{bibliography}

\end{document}